\pdfoutput=1

\documentclass[11pt]{article}

\usepackage[preprint]{acl}

\usepackage{times}
\usepackage{latexsym}

\usepackage[T1]{fontenc}

\usepackage[utf8]{inputenc}

\usepackage{microtype}

\usepackage{inconsolata}

\usepackage{graphicx,xcolor}
\usepackage[fleqn]{amsmath}
\usepackage{newtxtext}
\usepackage[varg]{newtxmath}
\usepackage[utf8]{inputenc}
\usepackage{CJK}

\usepackage[whole]{bxcjkjatype}

\usepackage{caption}
\usepackage{subcaption}
\usepackage{comment}
\usepackage{pmboxdraw}
\DeclareUnicodeCharacter{2581}{\pmboxdrawuni{2581}}
\usepackage{url}

\title{NAIST Simultaneous Speech Translation System for IWSLT 2024}

\author{Yuka Ko \quad Ryo Fukuda \quad Yuta Nishikawa \quad Yasumasa Kano \quad Tomoya Yanagita\\
\textbf{Kosuke Doi \quad Mana Makinae \quad Haotian Tan \quad Makoto Sakai}\\
\textbf{Sakriani Sakti \quad Katsuhito Sudoh \quad Satoshi Nakamura}\\
  Nara Institute of Science and Technology, Japan \\
  \texttt{ko.yuka.kp2@is.naist.jp}
  }

\begin{document}
\maketitle
\begin{abstract}
This paper describes NAIST's submission to the simultaneous track of the IWSLT 2024 Evaluation Campaign:
English-to-\{German, Japanese, Chinese\} speech-to-text translation and English-to-Japanese speech-to-speech translation.
We develop a multilingual end-to-end speech-to-text translation model combining two pre-trained language models, HuBERT and mBART.
We trained this model with two decoding policies, Local Agreement (LA) and AlignAtt. 
The submitted models employ the LA policy because it outperformed the AlignAtt policy in previous models. 
Our speech-to-speech translation method is a cascade of the above speech-to-text model and an incremental text-to-speech (TTS) module that incorporates a phoneme estimation model, a parallel acoustic model, and a parallel WaveGAN vocoder.
We improved our incremental TTS by applying the Transformer architecture with the AlignAtt policy for the estimation model.
The results show that our upgraded TTS module contributed to improving the system performance. 
\end{abstract}

\section{Introduction}
This paper presents NAIST's simultaneous speech translation (SimulST) systems for
the English-to-\{German, Japanese, Chinese\} speech-to-text track and the English-to-Japanese speech-to-speech track within the simultaneous track of the IWSLT 2024 Evaluation Campaign.

Simultaneous translation involves generating translations incrementally based on partial input, and it requires interpreters who can provide accurate and fluent translations while minimizing delay.

Early SimulST systems are based on a cascade of automatic speech recognition (ASR) and machine translation modules (\emph{e.g.,} \citealp{fugen2007simultaneous,bangalore-etal-2012-real,yarmohammadi-etal-2013-incremental,oda2014optimizing,arivazhagan2020re}), but they suffer from error propagation and added latency imposed by the ASR module.
Recently, an end-to-end approach has become popular \cite{agrawal-etal-2023-findings}, and this approach has been demonstrated to achieve a better quality-latency trade-off.

Conventional end-to-end SimulST models have employed training strategies and architectures designed for a simultaneous setting.
However, that approach not only requires additional effort in system development but also results in high computational costs.
To alleviate such problems, \citet{papi-etal-2022-simultaneous} proposed a single model trained on offline translation data for the simultaneous scenario.
Applying a simultaneous decoding policy to an offline speech translation (ST) model in SimulST inference enables the model to generate outputs similar to simultaneous translation.
Furthermore, a decoding policy determines whether to generate partial output or wait for more input.

Using an offline ST model with a simultaneous decoding policy has become popular because no specific task adaptation is required for a SimulST task.
Among several simultaneous decoding policies \cite{cho2016canneural,dalvi-etal-2018-incremental,ma-etal-2019-stacl,ma-etal-2020-simulmt,nguyen2021empirical}, Local Agreement (LA) \cite{liu2020lowlatency} is widely used and won the SimulST task at the IWSLT 2022 Evaluation Campaign \cite{anastasopoulos-etal-2022-findings}.
The LA policy extracts the longest common prefixes from the $n$ consecutive chunks as stable hypotheses.
However, it requires a long computation time to obtain the longest common prefix. 

Since simultaneous translation requires \emph{real-time} translation, a policy that runs fast is desirable.
\citet{papi23_interspeech} proposed a decoding policy called AlignAtt, which takes the alignments of the source and target tokens using cross attention information.
Under computation-aware settings, \citet{papi23_interspeech} have shown that AlignAtt can generate translations with lower latency compared to the LA policy, and it is capable of reaching a latency of 2 sec or less. 

For the IWSLT 2024 Evaluation Campaign, we developed two types of speech-to-text translation models with different decoding policies and compared them.
One is based on LA and the other on AlignAtt. 
The LA-based model demonstrates better quality than the AlignAtt-based one within the given latency constraints, while the AlignAtt policy works better in a low-latency region in computation-aware settings.

For the English-to-Japanese speech-to-speech track, we developed a cascade of the above SimulST model and an incremental text-to-speech module using a phoneme and prosodic symbol estimation model, a parallel acoustic model, and a parallel WaveGAN vocoder.
In last year's submission, our speech-to-speech translation method suffered from the quality of the synthesized speech and possible ASR errors \cite{fukuda-etal-2023-naist}.
The authors reported that the character error rate of the NAIST 2023 speech-to-speech translation output exceeded that of the SimulST text output by over 28\%.
Therefore, we upgraded our TTS module by incorporating Transformer architecture and AlignAtt in the estimation model.

\section{System Architecture}
This section describes the architecture of our SimulST systems.
First, we explain the decoding policies used for our translation modules.
Then, we present the details of our simultaneous speech-to-text and speech-to-speech translation methods.

\subsection{Decoding Policies}
\label{method:decoding_policy}
\subsubsection{Local Agreement}
\label{method:local_agreement}
\citet{liu2020lowlatency} introduced the concept of Local Agreement to find a stable prefix translation hypothesis in simultaneous translation scenarios where inputs are processed in fixed-length chunks.
This method assesses the stability of a hypothesis at step $t$ by comparing it with the hypothesis at step $t+1$, thus determining the agreeing prefix (i.e., the longest common prefix) between them.
The underlying principle is that the translation outputs with consistent agreeing prefixes, as the input prefixes increase, are likely to be reliable.
Building upon this idea, \citet{polak-etal-2022-cuni} extended it to encompass agreement among prefixes over $n$ consecutive steps (LA-$n$), with their experiments showing that $n=2$ performs effectively in the context of SimulST.
Based on these findings, we employed LA-2 as a SimulST policy and adjusted the input chunk length (in milliseconds) to manage the trade-off between quality and latency.

\subsubsection{AlignAtt}
Papi et al. \cite{papi23_interspeech} proposed AlignAtt, a method that leverages encoder-decoder attention information in Transformer to establish alignment between source and target tokens during inference. 
According to the AlignAtt policy, if a target token aligns with tokens beyond the last $f$ tokens of the source speech, it implies that adequate information has been provided to generate that token. 
Consequently, if a target token aligns solely with the last $f$ tokens from the source, generation is paused to await additional speech input.
In our implementation, we use cross attention from the decoder to the length adapter for AlignAtt.

\subsection{Simultaneous Speech-to-Text Translation}
\label{method:speech_to_text}
Our speech-to-text SimulST system uses multilingual offline speech translation models for the prefix-to-prefix translation required for SimulST. 
These models are based on large-scale pre-trained speech and text models adopting Hidden-Unit BERT (HuBERT) \cite{hsu2021hubert} and mBART50 \cite{tang2020multilingual}, following \citet{polak-etal-2022-cuni}. 
We initialized our ST models with the HuBERT speech encoder and the mBART50 text decoder, which were fine-tuned using English ASR data and multilingual MT data, respectively. 
In addition, we applied Inter-connection \cite{nishikawa23_interspeech} for the concatenated ST model. 
Inter-connection is a method that aggregates the information from each layer of a pre-trained speech model with weighted sums and then passes it into the decoder by connecting the intermediate layer of the speech encoder and the text decoder. 
We also fine-tuned the multilingual ST model using bilingual prefix pairs in English-to-\{German, Japanese, Chinese\} extracted using \emph{Bilingual Prefix Alignment} \cite{kano-etal-2022-simultaneous}. 
Bilingual Prefix Alignment is a method used to generate augmented prefix-to-prefix data based on a pre-trained offline model, and the SimulST model fine-tuned on those data will generate high quality output in a low-latency range compared to a model trained solely on offline data. 
After training these models, we applied the decoding policies in Section \ref{method:decoding_policy} to the ST model for controlling latency ranges. 

\subsection{Simultaneous Speech-to-Speech Translation}
\label{method:speech_to_speech}
Our English-to-Japanese speech-to-speech simultaneous translation is a cascade of the speech-to-text translation model (Section~\ref{method:speech_to_text}) and the incremental TTS module. 
In decoding steps, prefixes generated from the translation model are passed to the TTS module incrementally.
Then, the TTS module judges whether to wait for more inputs or generate a partial hypothesis.

\subsubsection{Incremental Text-to-Speech Synthesis}
\label{method:incremental_tts}
Incremental TTS consists of three modules: a phoneme estimator with a prosodic symbol for the Japanese language, an acoustic feature predictor, and a neural vocoder. 

The phoneme estimator predicts the phonemes of SimulST outputs and prosodic symbols in the Japanese language in parallel using a Transformer model.
This module uses three prosodic symbols to represent rising and falling pitches, and phrase boundary.
It works simultaneously with the input based on the AlignAtt policy using models trained in full-sentence conditions.
In TTS, it is assumed that there is monotonicity between input and output sequences, so there is little need to make delayed decisions and reorder, as is the case in LA. 
Therefore, we applied AlignAtt to TTS in this study. 
We modified the original Transformer architecture by adding two embedding input layers and two linear output layers to its decoder.
The self-attention mask was applied to both the encoder and the decoder sides because the subsequent sequences should not be used in the inference time in the incremental condition.

The acoustic feature predictor predicts acoustic features from the phonemes and prosodic symbols mentioned above, and then the neural vocoder synthesizes speech in parallel.
Its acoustic model is based on FastPitch \cite{9413889} with an additional adapter as an average phoneme power predictor.
Its encoder uses two independent embedding layers for phoneme and prosodic sequences and concatenates their embedding vectors into a single sequence as the input to the Transformer model.
Fastpitch estimates an acoustic feature sequence with predicted duration, pitch, and power in parallel.
Parallel WaveGAN synthesizes a speech waveform for the given acoustic features and noise sequences.

\section{Experimental Setting}
\subsection{Data}
\label{method:data}
\subsubsection{Simultaneous Speech-to-Text Translation}
We trained our multilingual ST model on MuST-C v2.0 \cite{di-gangi-etal-2019-must} and CoVoST-2 \cite{wang-etal-2020-covost} for all language pairs: English-to-German (En-De), English-to-Japanese (En-Ja), and English-to-Chinese (En-Zh). 
For the En-De setting, we also used MuST-C v1.0, Europarl-ST \cite{Europarl2020}, and TED-LIUM \cite{Rousseau2012TEDLIUMAA}.
In our training data, the development and test portions of CoVoST-2 and Europarl-ST were also included.
We used the MuST-C v2.0 tst-COMMON data as the evaluation data. 
We tokenized all of the text data in the corpora using a multilingual SentencePiece tokenizer with 250,000 subword units, distributed with the mBART50 model. 

For the En-Ja setting, we trained a model that applied a data filtering approach on the prefix translation pairs for the Bilingual Prefix Alignment data. 
We empirically set the ratio of the number of samples in the input speech to the number of tokens in the output at 4000.  
Any utterance exceeding the maximum ratio was excluded from the training data. 
In order to prevent discrepancies in sentence structure and word order between the source and target languages in fine-tuned models and thus avoid favoring shorter output. 

\subsubsection{Incremental Text-to-Speech Synthesis}
We used the JSUT corpus \cite{sonobe2017jsut} for training our FastPitch and Parallel WaveGAN. 
The numbers of sentences in the training, development, and test data were 7196, 250, and 250, respectively. 
For JSUT labels, we used the open-source repository \footnote{\url{https://github.com/r9y9/jsut-lab}}. 
We used the Balanced Corpus of Contemporary Written Japanese \cite{maekawa2014balanced} for training the phoneme and prosodic symbol estimation model.  
These symbols were obtained from the text using Open~Jtalk\footnote{\url{https://open-jtalk.sourceforge.net}} for training the estimation system. 
The same algorithm converted these symbols \cite{kurihara2021prosodic}, and symbols were separated into two sequences by adding blank tokens in prosodic symbols. 
The training, development, and test data were approximately 1.4 M, 10 K, and 10 K sentences, respectively. 
We also used the training portion of MuST-C as additional training data.  

\subsection{Simultaneous Speech-to-Text Translation}
\label{ssec:s2t-parameters}
We developed an end-to-end speech-to-text model by initializing it with two pre-trained models: HuBERT for the speech encoder and mBART50 for the text decoder. 
Furthermore, the encoder and decoder are interconnected via Inter-connection \cite{nishikawa23_interspeech} and a length adapter \cite{tsiamas-etal-2022-pretrained}. 
Speech input is provided as waveforms sampled at a rate of 16 kHz, which are then normalized to have zero mean and unit variance. 

We applied checkpoint averaging to the offline SimulST model.
During checkpoint averaging, model checkpoints were saved every 1000 training steps, and the averaged parameter values from the five best models, based on loss in the development data, were selected for the final model.

Subsequently, one epoch of fine-tuning was conducted on the training data, focusing solely on prefix alignment pairs in MuST-C v2. 
For this fine-tuning stage, the learning rate was reduced to $2.5\times10^{-5}$, using translation pairs obtained via Bilingual Prefix Alignment. 

For our SimulST strategies, we implemented both Local Agreement and AlignAtt policies. 
Specifically, we used Local Agreement with $n=2$ (LA-$2$). 
To adaptively control the quality-latency trade-off, we varied the chunk size from 200 to 1000 ms. 
During hypothesis generation for input chunks, a beam search with a beam size of five was employed. 
For the AlignAtt policy, we set the chunk size to 800 ms.
In AlignAtt, the parameter $f$ directly governs the model's latency: smaller values of $f$ imply that fewer frames are considered inaccessible by the model, thereby reducing the likelihood of the stopping condition being met and the resulting lower latency occurring. 
To adjust the quality-latency trade-off, we varied the parameter $f$ from 1 to 12. 
See Appendix~\ref{sec:s2t_parameter} for the detailed parameters of the speech-to-text model. 

\begin{table*}[t]
\centering
\caption{Results of submitted speech-to-text systems on MuST-C v2 tst-COMMON}
\begin{tabular}{lr|rrrrrr}
\hline
Language pair & Chunk size & BLEU & LAAL & AL & AP & DAL & ATD \\ \hline\hline
En-De & 960 ms & 29.978 & 2193.352 & 1973.799 & 0.846 & 2863.481 & 1887.436 \\
En-Ja & 835 ms & 15.329 & 2269.591 & 1868.759& 0.893 & 2878.447 & 541.729 \\
En-Zh & 910 ms & 22.300 & 2245.997 & 1959.588 & 0.839 & 2811.262 & 897.994 \\ \hline
\end{tabular}
\label{tab:system}
\end{table*}

\begin{table}[t]
\centering
\caption{Results of offline ST in submitted speech-to-text systems on MuST-C v2 tst-COMMON}
\begin{tabular}{lr}
\hline
Language pair  & BLEU \\ \hline\hline
En-De  & 31.00 \\
En-Ja  & 15.98 \\
En-Zh  & 24.98 \\ \hline
\end{tabular}
\label{tab:system_offline}
\end{table}

\subsection{Simulaneous Speech-to-Speech Translation}
\label{ssec:simul-speech-to-speech-system}
Our simultaneous speech-to-speech system was a cascade of the speech-to-text translation module and the incremental TTS module.
The parameter settings for the translation module were the same as those for the speech-to-text model, as described in Section~\ref{ssec:s2t-parameters}

\subsubsection{Incremental Text-to-Speech Synthesis}
The incremental TTS is composed of three modules: a phoneme estimator with a prosodic symbol for the Japanese language, an acoustic feature predictor, and a neural vocoder. 

For the phoneme estimator, the input vocabulary size was set to 21001.
The output vocabulary was set to 40  for phoneme and 4 for prosodic symbols.
The parameter of the AlignAtt policy $f$ was set to 1 in the phoneme and prosodic symbol estimation modules.
See Appendix~\ref{sec:tts-parameter} for the detailed parameters of the TTS model.

Speech was downsampled from 48 kHz to 22.05 kHz, and an 80-dimensional Mel spectrum was used for the acoustic features.
The size of the Fourier transform, frameshift length, window length, and window function were 2048, 10 ms, 50 ms, and Hann window, respectively. 

Our acoustic feature predictor mostly followed FastPitch structures, and the power predictor was added behind the pitch predictor. 

For the neural vocoder, experimental conditions for Parallel WaveGAN were the same as in the original paper, except for the parameters related to acoustic features and speech.

\begin{figure*}[t]
    \begin{tabular}{ccc}
      \centering
      \begin{minipage}[t]{0.3\textwidth}
        \centering
        \includegraphics[width=\linewidth]{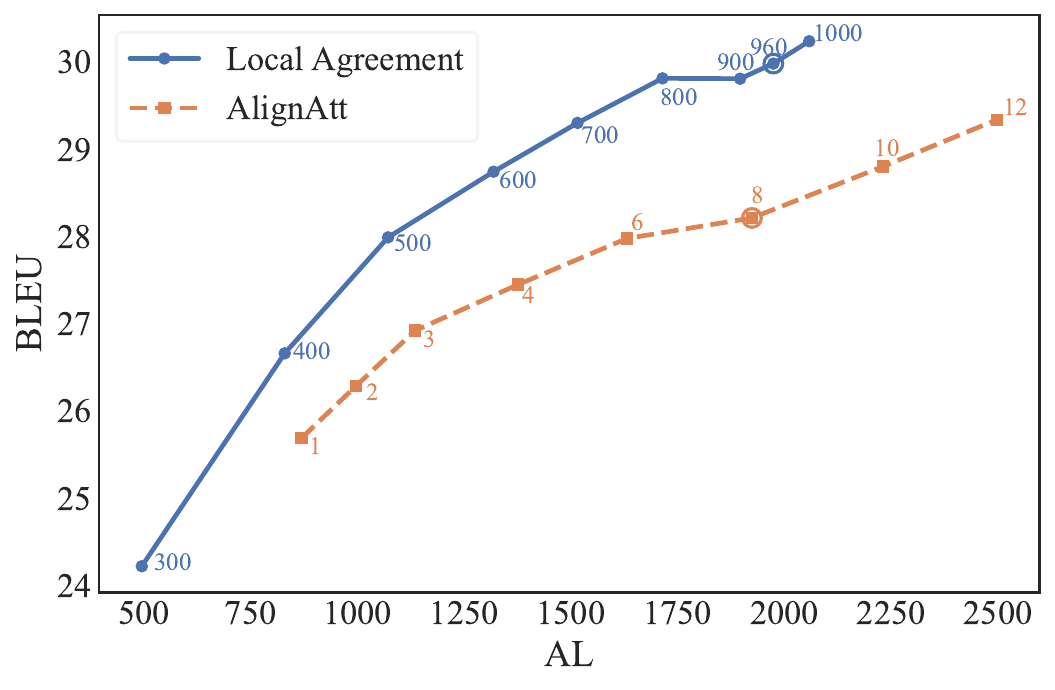}
        \subcaption{BLEU and AL in En-De}
      \end{minipage} &
      \begin{minipage}[t]{0.3\textwidth}
        \centering
        \includegraphics[width=\linewidth]{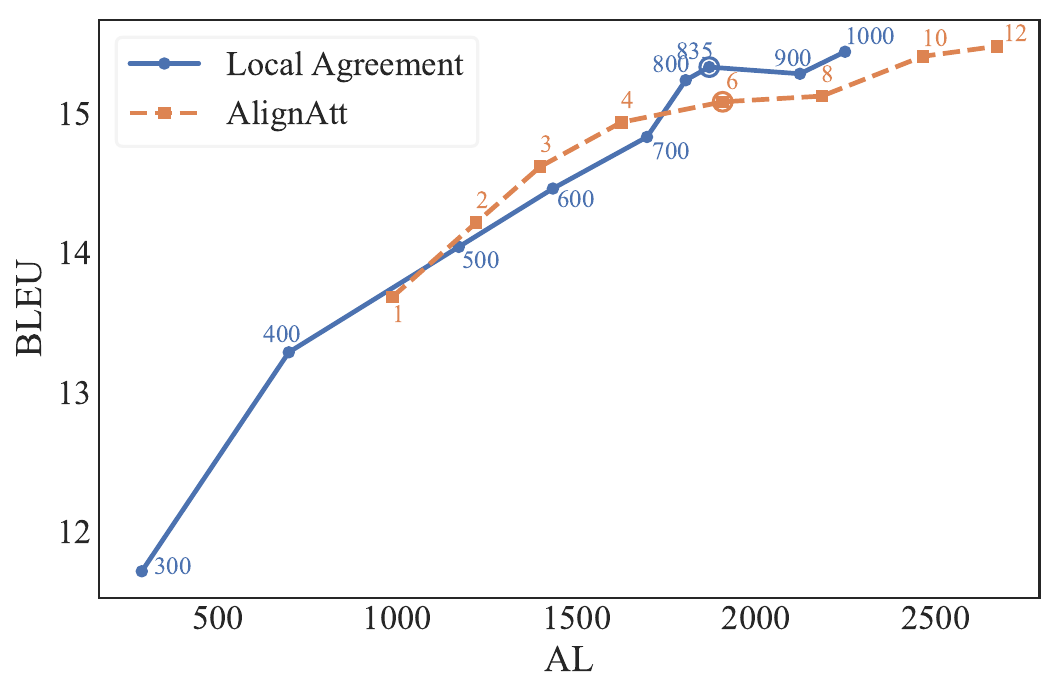}
        \subcaption{BLEU and AL in En-Ja}
      \end{minipage} &
      \begin{minipage}[t]{0.3\textwidth}
        \centering
        \includegraphics[width=\linewidth]{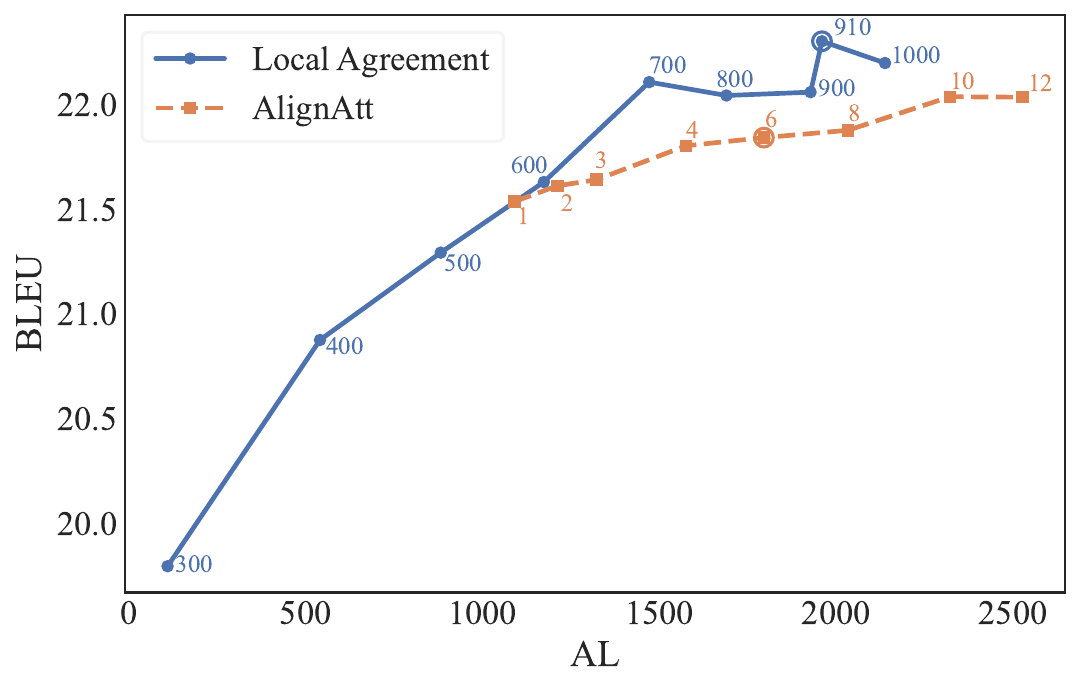}
        \subcaption{BLEU and AL in En-Zh}
      \end{minipage} \\\\
    \end{tabular}
    \caption{Results of \textbf{Local Agreement} and \textbf{AlignAtt} policies with AL on the speech-to-text systems. Circled dot in LA graph indicates our submitted system. Circled dot in AlignAtt graph indicates the best model satisfying the task requirement of IWSLT 2024 Shared Task.}
    \label{fig:s2t_nca}
\vspace{1cm}
    \begin{tabular}{ccc}
      \centering
      \begin{minipage}[t]{0.3\textwidth}
        \centering
        \includegraphics[width=\linewidth]{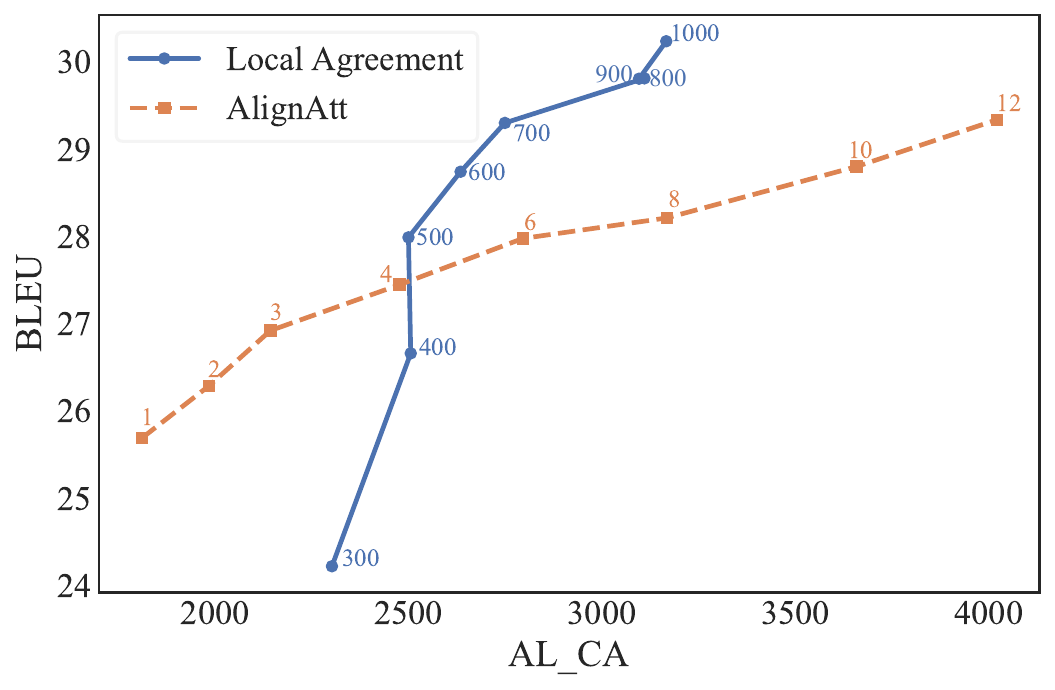}
        \subcaption{BLEU and AL\_CA in En-De}
      \end{minipage} &
      \begin{minipage}[t]{0.3\textwidth}
        \centering
        \includegraphics[width=\linewidth]{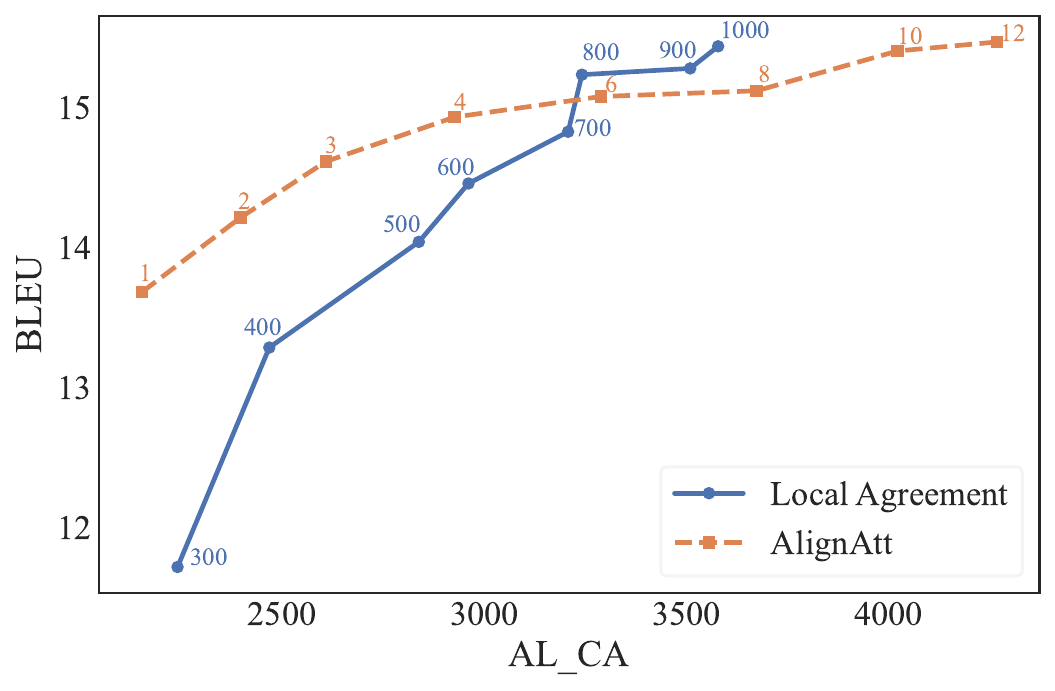}
        \subcaption{BLEU and AL\_CA in En-Ja}
      \end{minipage} &
      \begin{minipage}[t]{0.3\textwidth}
        \centering
        \includegraphics[width=\linewidth]{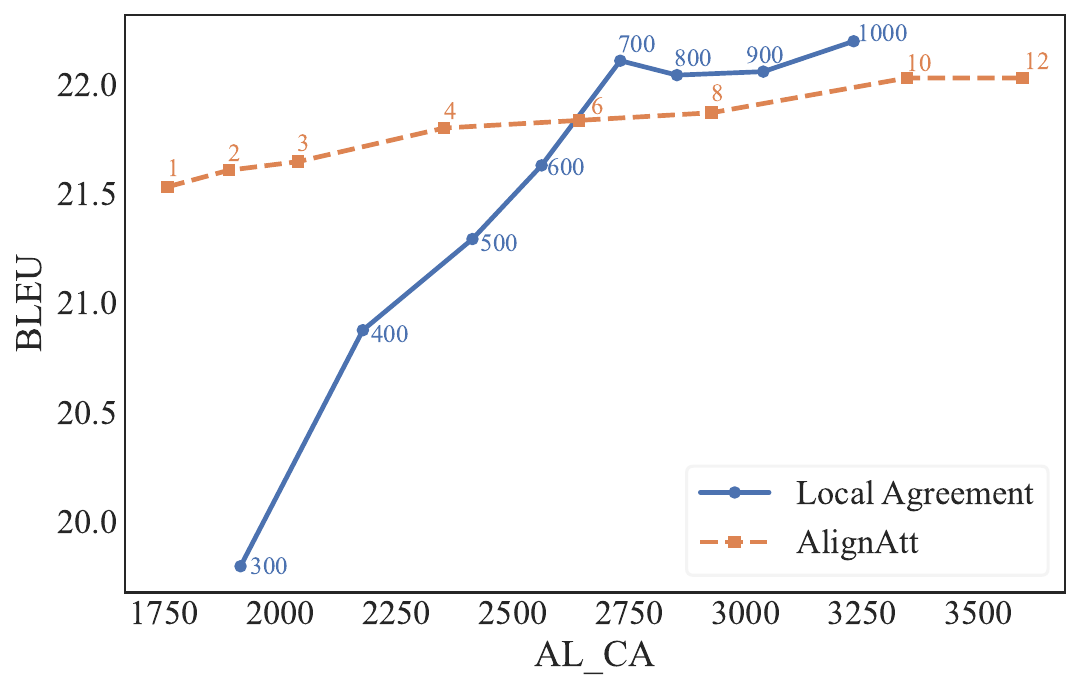}
        \subcaption{BLEU and AL\_CA in En-Zh}
      \end{minipage} \\\\
    \end{tabular}
    \caption{Results of \textbf{Local Agreement} and \textbf{AlignAtt} policies with AL\_CA on the the speech-to-text systems}
    \label{fig:s2t_ca}
  \end{figure*}

\subsection{Evaluation}
We assessed our systems using the SimulEval \cite{ma-etal-2020-simuleval} toolkit\footnote{\url{https://github.com/facebookresearch/SimulEval}} and evaluated the translation quality of the SimulST systems using BLEU with sacreBLEU\footnote{\url{https://github.com/mjpost/sacrebleu}}.
We also measured translation latency by the following metrics:
\begin{itemize}
    \setlength{\itemsep}{-3pt}
    \item Average Lagging (AL) \cite{ma-etal-2019-stacl}
    \item Length Adaptive Average Lagging (LAAL) \cite{papi-etal-2022-generation}
    \item Average Token Delay (ATD) \cite{kano2024atd}
    \item Average Proportion (AP) \cite{cho2016canneural}
    \item Differentiable Average Lagging (DAL) \cite{cherry2019thinking}
\end{itemize}

For the SimulS2S system, translation quality was evaluated using BLEU scores obtained after transcribing the output speech with Whisper \cite{radford2022robust} (ASR\_BLEU). 
Translation latency was evaluated using ATD along with Start\_Offset and End\_Offset \cite{agrawal-etal-2023-findings}. 

AL is a widely used latency metric for both text-to-text and speech-to-text simultaneous translation. 
However, while AL focuses on the time translation begins, it does not adequately consider the time each input chunk's translation ends. 
In scenarios where speech segments are generated sequentially, as in speech-to-speech translation, the translation output may be delayed if the preceding outputs occupy the speech output channel. 
Consequently, AL may not be suitable for evaluating the latency of speech-to-speech simultaneous translation. 
Instead, we employ ATD, which includes delays caused by output in the latency calculation. 
ATD computes delays by calculating the average time difference between each source token and its corresponding target token. 
In the SimulEval setup, assuming each word requires 300 ms to be spoken, both the input and output speech are segmented into 300-ms intervals, treating these segments as tokens for ATD calculations. 

\section{Experiment Results}

\subsection{Simultaneous Speech-to-Text System}
We chose one submission for each language direction, ensuring that the settings met the task requirement of $AL \leq 2\ \textrm{sec}$.
The submission model is based on the LA policy, since it outperformed the AlignAtt policy used in earlier models. 

\subsubsection{NAIST 2023 model vs. 2024 model}
Table~\ref{tab:system} shows the results of the submitted speech-to-text systems evaluated on MuST-C v2 tst-COMMON.
Although the system architecture of our submitted models was the same as that of last year's models, the chunk size settings were different in every language pair.
Using different chunk size settings slightly improved the BLEU scores in every language pair (see Appendix~\ref{sec:appendix_a} for the scores for our 2023 submission).
We also show the results of the offline ST in submitted speech-to-text systems on MuST-C v2 tst-COMMON in Table~\ref{tab:system_offline}. 

\subsubsection{Local Agreement vs. AlignAtt Policies} 
Figure~\ref{fig:s2t_nca} shows BLEU and AL trade-offs in non-computation-aware conditions.
When comparing the results of the LA and AlignAtt policies, there was little difference observed in En-Ja (Figure~\ref{fig:s2t_nca} (b)), while there were relatively large gaps in BLEU in En-De and En-Zh, especially in the high latency region (Figures~\ref{fig:s2t_nca} (a) and (c)).

Figure~\ref{fig:s2t_ca} shows the BLEU and AL trade-offs in computation-aware conditions. 
In all language pairs, the AlignAtt policy was better in the low-latency region, while the LA policy was better in the high-latency regions. 

\begin{table*}[t]
\centering
\caption{Results of submitted SimulS2S system on the MuST-C v2 tst-COMMON}
\begin{tabular}{lc|rrrr}
\hline
System & Chunk size & ASR\_BLEU & Start\_Offset & End\_Offset & ATD \\ \hline\hline
LA (NAIST 2023) & 650 ms & 9.873 & 2495.010 & 4134.752 & 3278.809 \\
LA (NAIST 2024) & 950 ms & 12.082 & 2425.485 & 3745.743 & 3792.405 \\
AlignAtt & 800 ms ($f$=6) & 11.650 & 2493.908 & 3505.377 & 3682.920 \\ \hline

\end{tabular}
\label{tab:system_s2s}
\end{table*}

\begin{figure*}[t]
    \begin{tabular}{ccc}
      \centering
      \begin{minipage}[t]{0.3\textwidth}
        \centering
        \includegraphics[width=\linewidth]{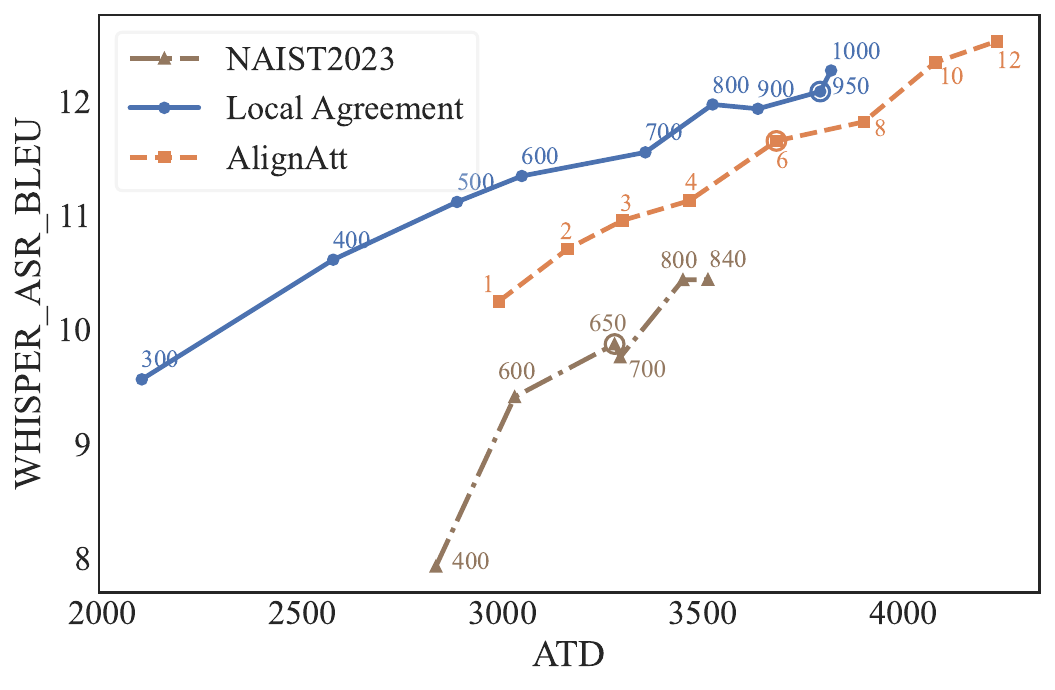}
        \subcaption{BLEU and ATD}
      \end{minipage} &
      \begin{minipage}[t]{0.3\textwidth}
        \centering
        \includegraphics[width=\linewidth]{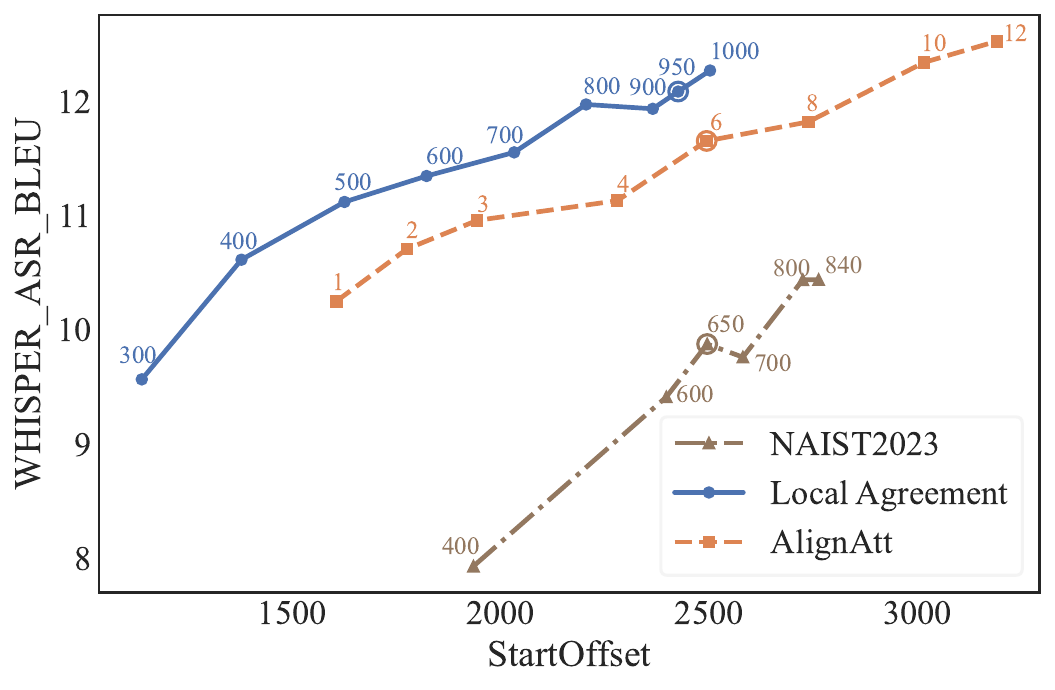}
        \subcaption{BLEU and Start\_Offset}
      \end{minipage} &
      \begin{minipage}[t]{0.3\textwidth}
        \centering
        \includegraphics[width=\linewidth]{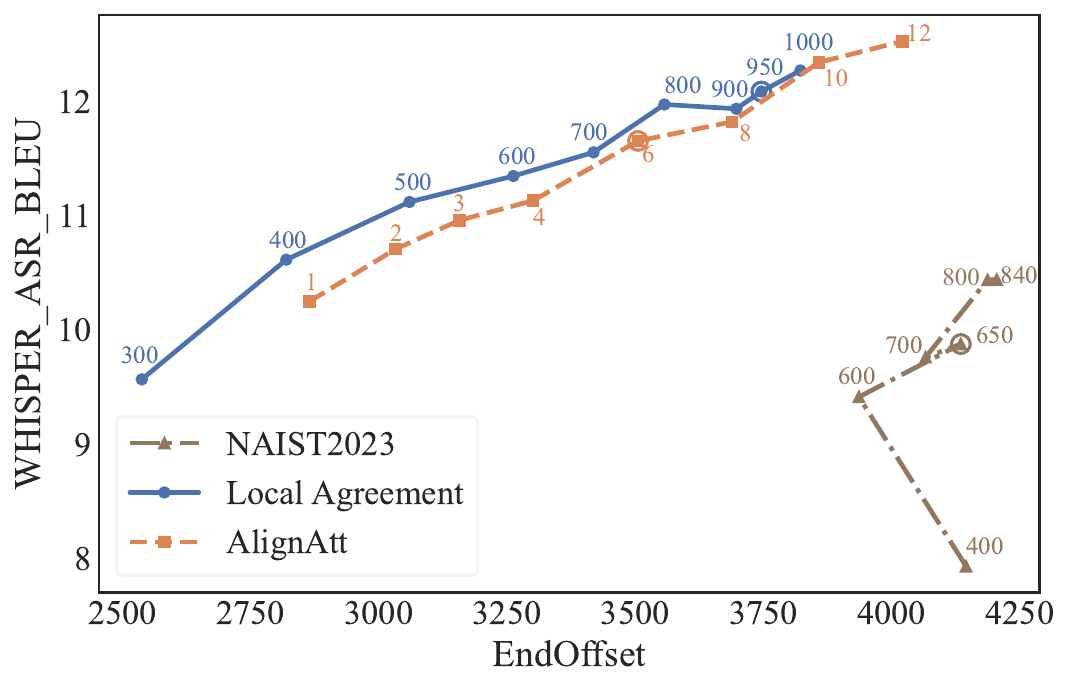}
        \subcaption{BLEU and End\_Offset}
      \end{minipage} \\\\
    \end{tabular}
    \caption{Results of \textbf{Local Agreement} and \textbf{AlignAtt} policies with ATD, Start\_Offset, and End\_Offset on speech-to-speech systems. Circled dot in LA graph indicates submitted system. Circled dot in AlignAtt graph indicates the best model satisfying the task requirement of IWSLT 2024 Shared Task.}
    \label{fig:s2s_nca}
\vspace{1cm}
    \begin{tabular}{ccc}
      \centering
      \begin{minipage}[t]{0.3\textwidth}
        \centering
        \includegraphics[width=\linewidth]{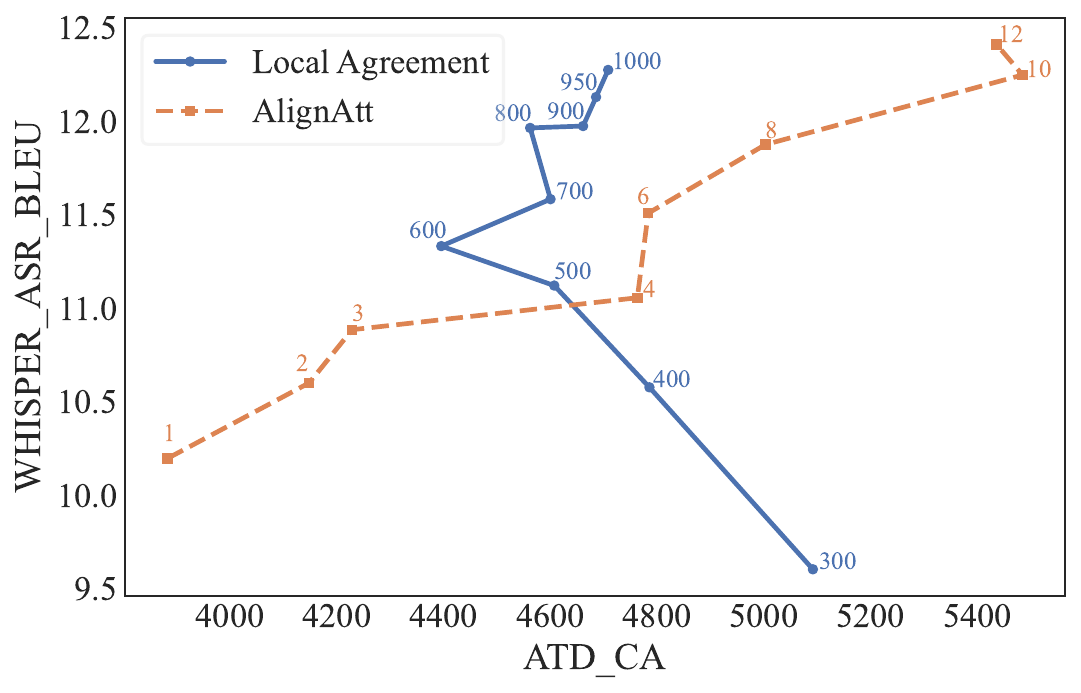}
        \subcaption{BLEU and ATD\_CA}\label{fig:s2s_atd_ca}
      \end{minipage} &
      \begin{minipage}[t]{0.3\textwidth}
        \centering
        \includegraphics[width=\linewidth]{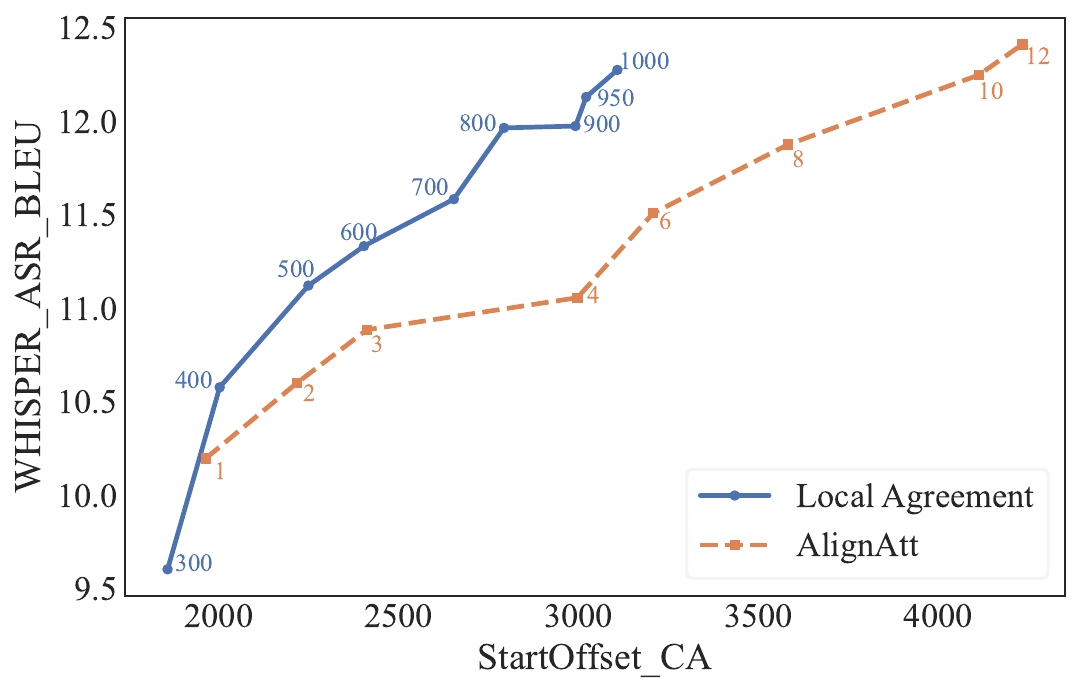}
        \subcaption{BLEU and Start\_Offset\_CA}
      \end{minipage} &
      \begin{minipage}[t]{0.3\textwidth}
        \centering
        \includegraphics[width=\linewidth]{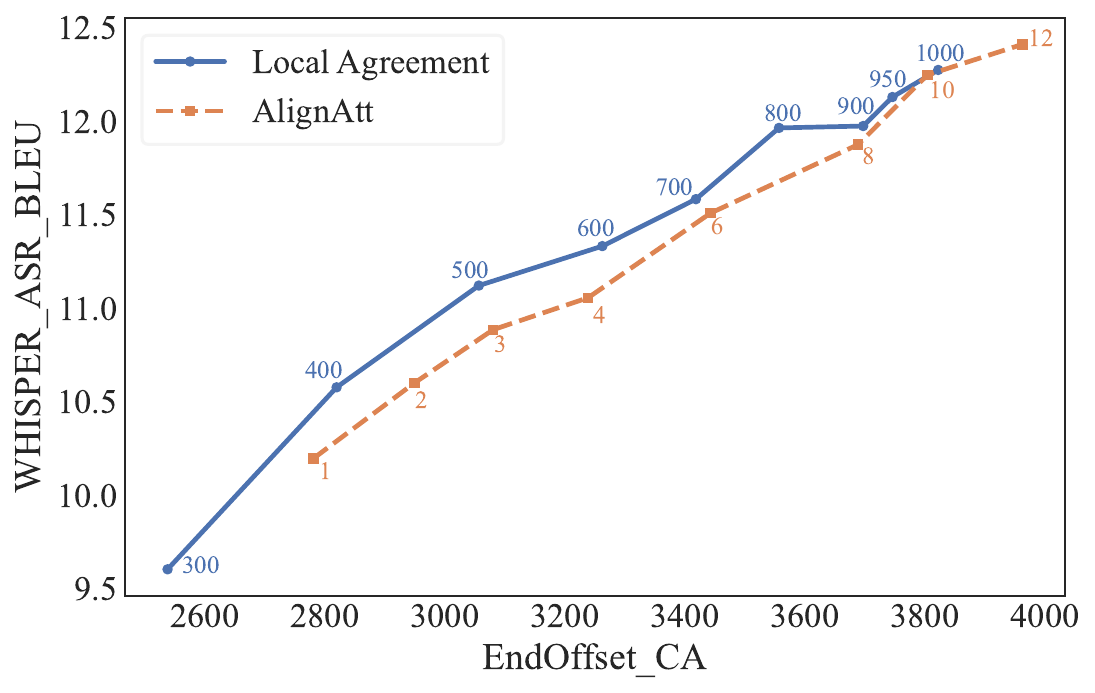}
        \subcaption{BLEU and End\_Offset\_CA}
      \end{minipage} \\\\
    \end{tabular}
    \caption{Results of \textbf{Local Agreement} and \textbf{AlignAtt} policies with ATD\_CA, Start\_Offset\_CA and End\_Offset\_CA on speech-to-speech translation systems.}
    \label{fig:s2s_ca}
  \end{figure*}

\subsubsection{Non-Computation-Aware vs. Computation-Aware Latency}
The quality-latency trade-off results differed significantly between the non-computation-aware and the computation-aware conditions.
The LA policy requires a relatively long computation time to obtain the longest common prefixes.
This is especially true when the source speech is divided into many small segments.
Therefore, the latency increases significantly when a small chunk size is set (see Figure~\ref{fig:s2t_ca}).

The main constraint of the IWSLT 2024 Shared Task (\emph{i.e.}, latency is measured in a non-computation-aware setting) may have been advantageous for the LA policy.
In fact our LA-based system outperformed our AlignAtt-based one.
However, in reality, the LA policy is time-consuming, and thus the AlignAtt policy may be better suited to practical applications. 

\subsection{Simultaneous Speech-to-Speech System} 
We submitted a model with the LA policy for the En-Ja speech-to-speech track.
We selected a model configured with a chunk size of 950 ms, which satisfies the task requirement Start\_Offset $\leq 2.5\ \textrm{sec}$.
Table~\ref{tab:system_s2s} shows the results of our speech-to-speech model (LA (NAIST 2024)).
We also developed a model with the AlignAtt policy, but the LA model achieved higher ASR\_BLEU than the AlignAtt model.
The quality-latency trade-offs in non-computation-aware and computation-aware conditions are shown in Figures~\ref{fig:s2s_nca} and \ref{fig:s2s_ca}. 

\subsubsection{NAIST 2023 model vs. 2024 model}
Our submitted model outperformed our last year's submission (LA (NAIST 2023)).
We compared our 2024 submission with the 2023 one to clarify what contributed to improving the score.
The significant difference between the two systems lies in the upgraded TTS, which has an estimation model based on Transformer architecture with the AlignAtt policy (see Section~\ref{ssec:simul-speech-to-speech-system}).

When comparing the output from the speech translation modules, there was little difference in BLEU scores between the two systems (2023 system: 14.93; 2024 system: 15.44)\footnote{The chunk size setting for the 2024 speech-to-speech system was different from that for the 2024 speech-to-text system.}. 
However, the performance of our 2024 system, which was measured by ASR\_BLEU, was more than 2 points higher than that of our 2023 system.
The results suggest that our new TTS contributed to the improved score.
We listened to samples of synthesized speech and observed that the outputs from the 2024 system tended to be more natural in accent and intonation compared to those from the 2023 system.

\subsubsection{Local Agreement vs. AlignAtt Policies}
We further compared the model with the LA policy (our 2024 submission) with the model with the AlignAtt policy.
Comparing the translation modules of the two systems, the difference in translation quality measured by BLEU was about 0.4 points (15.44 and 15.09 for the LA and the AlignAtt models, respectively).
This gap is almost the same as the gap in the evaluation scores of the speech-to-speech systems (ASR\_BLEU, see Table~\ref{tab:system_s2s}).

Although no difference was observed in BLEU scores, a comparison between the output from the speech-to-speech system (\emph{i.e.}, transcribed speech) with the output from the translation module suggests that the policy difference affects the TTS performance.
We extracted sentences that satisfy the following criteria: (1)~translations from the translation modules were identical between the two policies, but (2)~transcribed speeches were different between the two systems.

We computed BLEU scores using extracted sentences ($N$=414) while regarding the outputs from the translation modules as references.
The score for the LA policy was more than 2 points higher than that for the AlignAtt policy (67.79 and 65.46, respectively).
In addition, the transcribed speech for AlignAtt was shorter than that for LA (sys\_len: 6364 and 6464, respectively).
These results suggest that the manner of passing the translations to the TTS was different between the two decoding policies (\emph{e.g.}, timing) and affected the TTS performance.

\begin{figure*}[t]
    \centering
      \begin{minipage}[t]{\textwidth}
        \centering
        \includegraphics[width=\linewidth]{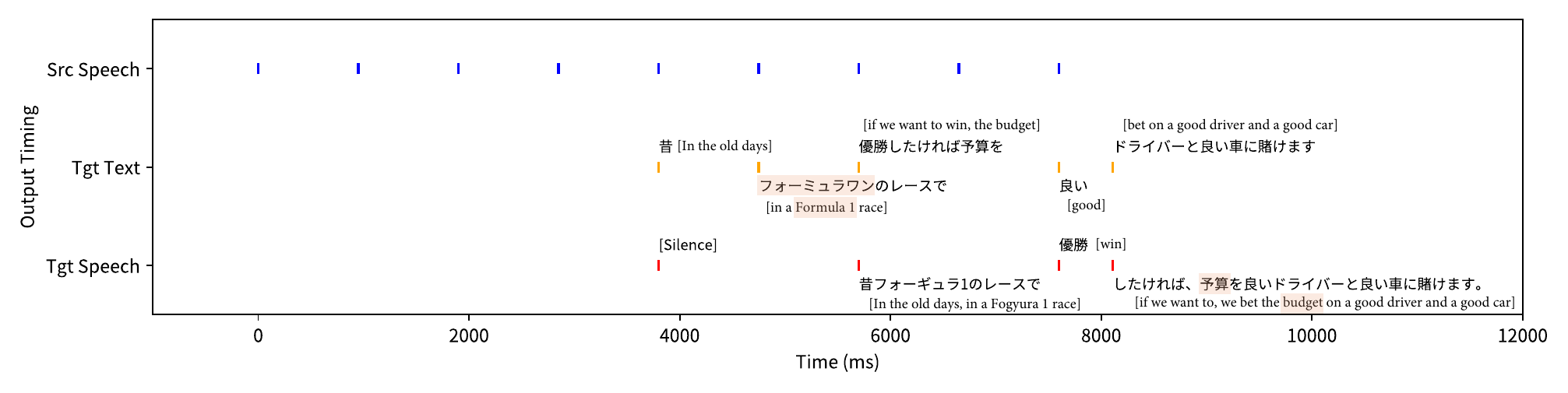}
        \subcaption{LA (chunk size = 950 ms)}
        \label{fig:la_diagram}
      \end{minipage}
      
      \begin{minipage}[t]{\textwidth}
        \centering
        \includegraphics[width=\linewidth]{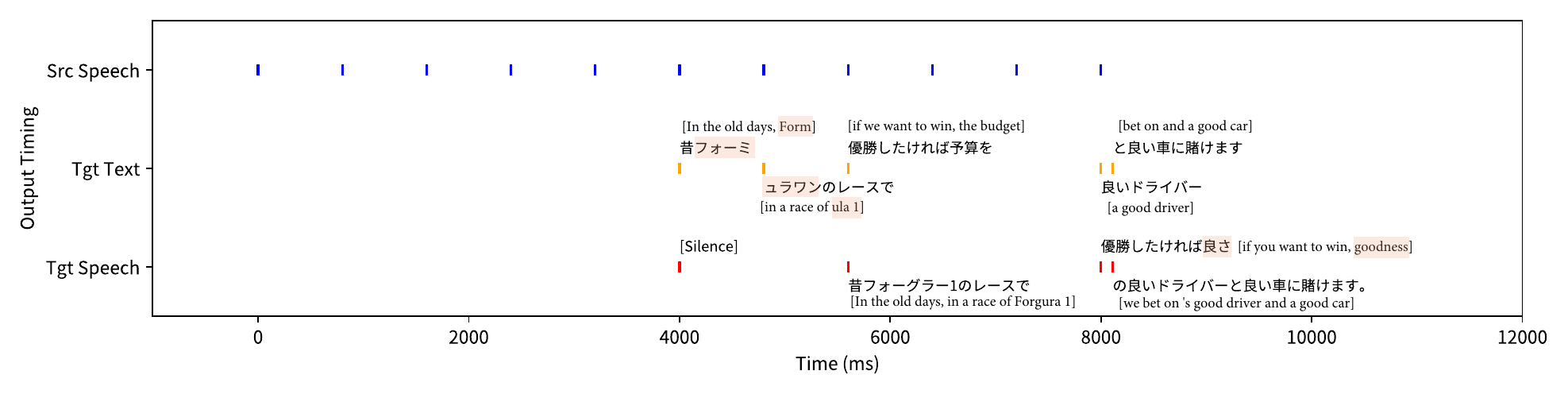}
        \subcaption{AlignAtt (chunk size = 800 ms, $f=6$)}
        \label{fig:alignatt_diagram}
      \end{minipage}
    \caption{Example of timing difference in passing translations to the TTS between the LA and AlignAtt policies. Translations generated by speech-to-text models were identical between the two policies, but outputs from the TTS module were different. }
    \label{fig:timing_diagram}
\end{figure*}

Our analysis suggests that the output from the LA policy was more suitable for our TTS than that from the AlignAtt policy because the LA policy generated longer partial output with more confident agreement.
On the other hand, the AlignAtt policy tended to generate prefixes whose boundaries did not correspond to meaningful units and sometimes divided a word in the middle of it.
Figure~\ref{fig:timing_diagram} shows an example of the timing difference in passing the translations to the TTS.
This figure compares the output prefixes generated from the translation modules with different decoding policies and the output prefixes generated from the TTS module along with the timing information.   
In this example, the translations generated from the translation modules are identical between the LA and AlignAtt policies.
However, the prefixes (see Tgt text in Figure~\ref{fig:timing_diagram}) and the timing when they were passed to the TTS module were different between the two decoding policies.
In this example, the LA policy tended to generate semantically coherent prefixes, which resulted in more successful output from the TTS module (see Tgt speech).
On the other hand, the AlignAtt policy divided the word ``フォーミュラワン [Formula 1]'' into two prefixes, ``フォーミ [Form]'' and ``ュラワン [ula 1].''
When the boundaries of the prefixes do not correspond to the meaning units or words are divided into prefixes, it might be difficult to capture the context of a sentence, which results in poor performance of the TTS module. 
In this example, the word ``予算 [budget]'' (pronounced as \emph{yosan}) was wrongly recognized as ``良さ [goodness]'' (pronounced as \emph{yosa}) in the system with the AlignAtt policy.
The results suggest that feeding stable prefixes to the TTS module is important in our speech-to-speech system.
Future study will involve making the AlignAtt policy generate more stable prefixes.

\subsubsection{Non-Computation-Aware vs. Computation-Aware Latency}
Figures~\ref{fig:s2s_nca} and \ref{fig:s2s_ca} show the results in non-computation-aware and computation-aware settings, respectively.
When the latency was measured by the Start\_Offset and the End\_Offset, there were no large differences between the results in non-computation-aware and computation-aware settings.
However, when latency was measured by ATD, the quality-latency trade-offs exhibited different trends in non-computation-aware and computation-aware settings.

Start\_Offset does not include computation time as a delay because Start\_Offset is measured only at the start of translation.
Therefore, Start\_Offset is not appropriate as the latency metric in computation-aware settings.
Moreover, Start\_Offset and End\_Offset measure the delay at a single point in the translation and does not consider the delays in the middle section of the translation.

In contrast, ATD measures the delay at multiple points and has a higher correlation with Ear-Voice Span, which is often used as a reference latency metric in human interpretation research \cite{kano2024atd}.
As the segments become smaller, the number of segments increases.
This increases the number of comparison processes at the inference of LA.
Therefore, the computation time becomes larger as the segment size becomes smaller and BLEU becomes lower in the low-latency range of LA, which is only shown in Figure~\ref{fig:s2s_atd_ca}.

In a computation-aware setting, we observed that the AlignAtt policy outperformed the LA policy in the low-latency region (Figure~\ref{fig:s2s_ca}).
In practical situations, the LA policy might be time-consuming for a speech-to-speech system.
One future direction would be improving the performance of a model with the AlignAtt policy. 

\section{Conclusions}
In this paper, we described our SimulST systems for the IWSLT 2024 Simultaneous Speech Translation task. 
Experimental results demonstrated the effectiveness of AlignAtt by comparison to Local Agreement in terms of computation-aware latency, especially in the low-latency range. 
Our speech-to-speech translation system also showed the effectiveness of applying AlignAtt to the TTS model and resulted in better performance compared to our IWSLT 2023 system. 
This time, our speech-to-text method used HuBERT with the mBART model, while our TTS method only used the Parallel WaveGAN vocoder. 
In the future, we will investigate other methods such as WavLM~\cite{chen2022wavlm} and Hi-Fi GAN~\cite{kong2020hifi}. 

\section*{Acknowledgements}
Part of this work was supported by JSPS KAKENHI Grant Number JP21H05054.


\appendix

\section{Speech-to-Text Parameter Settings}
\label{sec:s2t_parameter}
\vspace{-0.1cm}
The speech encoder was initialized with HuBERT-Large, comprising a feature extractor trained on 60 K hours of unlabeled speech data from Libri-Light \cite{librilight}, along with Transformer encoder layers. 
The feature extractor consists of seven convolutional layers with kernel sizes of (10, 3, 3, 3, 3, 2, 2), corresponding strides of (5, 2, 2, 2, 2, 2, 2), and 512 channels. The number of Transformer encoder layers is 24.
The text decoder was initialized using the decoder component of mBART50.
The decoder is composed of twelve Transformer layers, sharing an embedding layer and linear projection weights sized at 250,000. Each Transformer and feed-forward layer has dimensions of 1024 and 4096, respectively, with 16 attention heads. 
ReLU serves as the activation function, and layer normalization is applied before attention operations.
The length adapter is implemented as a three-layer convolutional network featuring 1024 channels, a stride of 2, and a Gated Linear Unit (GLU) activation function.
During training, each source audio was augmented \cite{wavaugment2020} prior to normalization, with a probability of 0.8.
Multilingual models were trained using all of the data with a maximum source length of 400,000 frames and a target length of 1024 tokens. 
To achieve a batch size of approximately 32 million tokens, we employ gradient accumulation and data-parallel computations.
We utilize the Adam optimizer with $\beta_1=0.99$, $\beta_2=0.98$, and a base learning rate of $2.5\times10^{-4}$. A tri-stage scheduler controls the learning rate, with warm-up, hold, and decay phases set to 0.15, 0.15, and 0.70, respectively. 
The initial and final learning rates are scaled to 0.01 compared to the base rate. 
Sentence averaging and gradient clipping of 20 are applied, along with a dropout probability of 0.1. 
Time masking is used for 10-length spans with a probability of 0.2, while channel masking is applied to 20-length spans with a probability of 0.1 in the output of the encoder's feature extractor. 
The loss function employed is cross-entropy with label smoothing of 20\% probability mass.

\section{Incremental Text-to-Speech Parameter Settings}
\label{sec:tts-parameter}
\vspace{-0.1cm}
For the phoneme estimator, each Transformer layer, head, dimension of head, or dimension of Transformer model was 2, 8, 64, or 512, respectively. 
The embedded size of the encoder was the same as the dimension of the Transformer, and the embedding sizes of the decoder were 128 dimensions for the prosodic symbols and 512 dimensions for the phonemes. 
The batch size for training was 256. 
We used Adam optimizer with a learning rate of 0.1, $\beta_1=0.99$, $\beta_2=0.99$, and $\epsilon=1e-8$. 
The warmup scheduler is the same as that of the original Transformer.
The size of the Fourier transform, frameshift length, window length, and window function were 2048, 10 ms, 50 ms, and Hann window, respectively. 
The changed settings were as follows: We used two embedding layers with a hidden size of 256, the hidden size in Transformer was 256, the number of heads was 2, the encoder and decoder had 4 layers, the first convolution layer in each FFT block in FastPitch had a kernel size of 3 and 256/1024 input/output channels, the second convolution layer in an FFT block had 1024/256 input/output channels with the same kernel size, the first convolution layer in each predictor had a kernel size of 3 and 256/256 input/output channels, and the second convolution layer in each predictor had 256/256 input/output channels with the same kernel size. 
We used Adam optimizer with a learning rate of 0.1, $\beta_1=0.99$, $\beta_2=0.99$, and $\epsilon=1e-9$. 
The batch size was 48. 
The schedule for the warmup followed FastPitch. 

\begin{table*}[t]
\centering
\caption{Results of submitted speech-to-text systems on MuST-C v2 tst-COMMON in IWSLT 2023}
\begin{tabular}{lr|rrrrrr}
\hline
Language pair & Chunk size & BLEU & LAAL & AL & AP & DAL & ATD \\ \hline\hline
En-De & 950 ms & 29.975 & 2172.927 & 1964.329 & 0.846 & 2856.738 & 1893.749 \\
En-Ja & 840 ms & 15.316 & 2290.716 & 1973.586 & 0.892 & 2889.950 & 547.752 \\
En-Zh & 700 ms & 22.105 & 1906.995 & 1471.287 & 0.821 & 2436.948 & 667.780 \\ 
\hline
\end{tabular}
\label{tab:system_2023_main_s2t}
\end{table*}

\section{NAIST 2023 Submission for Speech-to-Text}
\label{sec:appendix_a}
\vspace{-0.1cm}
Table \ref{tab:system_2023_main_s2t} shows the results for all chunk size settings for the En-De, En-Ja, and En-Zh models, respectively, used in our 2023 submission \cite{fukuda-etal-2023-naist}.

\end{document}